\documentclass[sn-mathphys,Numbered, iicol]{sn-jnl}

\usepackage{array}
\usepackage[labelfont=small,textfont=scriptsize]{subfig}
\usepackage{stfloats}
\usepackage{url}
\usepackage{verbatim}
\usepackage{bm}
\usepackage{balance}
\usepackage{dsfont}
\usepackage{lineno}
\usepackage{lipsum}
\usepackage{ulem}
\usepackage{booktabs}
\usepackage{natbib}
\usepackage{enumerate}

\setcitestyle{authoryear,open={(},close={)},semicolon} 

\usepackage{graphicx}%
\usepackage{multirow}%
\usepackage{amsmath,amssymb,amsfonts}%
\usepackage{amsthm}%
\usepackage{mathrsfs}%
\usepackage{xcolor}%
\usepackage{textcomp}%
\usepackage{manyfoot}%
\usepackage{algorithm}%
\usepackage{algorithmicx}%
\usepackage{algpseudocode}%
\usepackage{listings}%

\definecolor{wine}{rgb}{0.5069,0.0931,0.4}
\definecolor{vblue}{rgb}{0.388,0,0.969}
\definecolor{salmon}{rgb}{1,0.561,0.353}
\definecolor{dgreen}{rgb}{0.043,0.388,0.161}
\definecolor{lgreen}{rgb}{0.243,0.933,0.651}

\newcommand{\bd}[1]{\boldsymbol{#1}}
\newcommand{\s}[1]{\mathcal{#1}}

\newcommand{\dgreen}{\fcolorbox{black}{dgreen}{\rule{0pt}{2pt}\rule{2pt}{0pt}}}
\newcommand{\wine}{\fcolorbox{black}{wine}{\rule{0pt}{2pt}\rule{2pt}{0pt}}}
\newcommand{\lgreen}{\fcolorbox{black}{lgreen}{\rule{0pt}{2pt}\rule{2pt}{0pt}}}
\newcommand{\salmon}{\fcolorbox{black}{salmon}{\rule{0pt}{2pt}\rule{2pt}{0pt}}}
\newcommand{\vblue}{\fcolorbox{black}{vblue}{\rule{0pt}{2pt}\rule{2pt}{0pt}}}

\begin{document}
\title{Large-scale unsupervised spatio-temporal semantic analysis of vast regions from satellite images sequences}


\author[1,2]{\fnm{Carlos} \sur{Echegoyen}}\email{carlos.echegoyen@unavarra.es}

\author[3]{\fnm{Aritz} \sur{Pérez}}\email{aperez@bcamath.org}

\author[1,2]{\fnm{Guzmán} \sur{Santafé}}\email{guzman.santafe@unavarra.es}

\author[1,2]{\fnm{Unai} \sur{Pérez-Goya}}\email{unai.perez@unavarra.es}

\author[1,2]{\fnm{María Dolores} \sur{Ugarte}}\email{lola@unavarra.es}

\affil[1]{\orgdiv{Department of Statistics, Computer Science and Mathematics}, \orgname{Public University of Navarre}, \orgaddress{\city{Pamplona}, \postcode{31006}, \country{Spain}}}

\affil[2]{\orgdiv{INAMAT$^2$}, \orgname{Public University of Navarre}, \orgaddress{\city{Pamplona}, \postcode{31006}, \country{Spain}}}

\affil[3]{\orgname{Basque Center for Applied Mathematics}, \orgaddress{\city{Bilbao}, \postcode{48009}, \country{Spain}}}

\abstract{Temporal sequences of satellite images constitute a highly valuable and abundant resource for analyzing regions of interest. However, the automatic acquisition of knowledge on a large scale is a challenging task due to different factors such as the lack of precise labeled data, the definition and variability of the terrain entities, or the inherent complexity of the images and their fusion.
In this context, we present a fully unsupervised and general methodology to conduct spatio-temporal taxonomies of large regions from sequences of satellite images. Our approach relies on a combination of deep embeddings and time series clustering to capture the semantic properties of the ground and its evolution over time, providing a comprehensive understanding of the region of interest. The proposed method is enhanced by a novel procedure specifically devised to refine the embedding and exploit the underlying spatio-temporal patterns.
We use this methodology to conduct an in-depth analysis of a 220 km$^2$ region in northern Spain in different settings.
The results provide a broad and intuitive perspective of the land where large areas are connected in a compact and well-structured manner, mainly based on climatic, phytological, and hydrological factors.}



\maketitle

\section{Introduction}
Earth monitoring using satellite image analysis is nowadays essential for the identification, mapping, assessment, and monitoring of land use and land cover change.  This land monitoring throughout long periods of time is possible and cost-effective thanks to multi-spectral satellite images freely provided by satellite programs supported by public agencies, such as the Sentinel-2 program by the European Space Agency. Thus, public access to satellite imagery has favored the interest of a growing number of researchers in the analysis of satellite image time series (SITS). The vast data volume and the complexity of the SITS analysis have promoted the use of machine learning methods to obtain land use maps, land cover maps, crop classification, or harvest prediction \citep{Csillik2019, Qiao2021, Wambugu2021}.  However, obtaining labeled data may be problematic in SITS analysis since they are very expensive to produce and maintain. Therefore, semi-supervised and clustering methods are gaining more and more attention in this field \citep{Jean2019, Kalinicheva20, Chen2022b}.  
Moreover, when dealing with large regions, it is crucial to employ effective data partitioning strategies \citep{lin2020parallel,olasz2016new,wu2021recent}. Typically, satellite imagery is generated using a standardized overlapping grid to prevent image gaps and facilitate image fusion. However, this approach often results in a noticeable border effect during image analysis.


In recent years, the rapid development of convolutional neural networks (CNN) represents a revolution in the field of image analysis in general and in SITS data in particular \citep{Shutao19,Moskolai21}. CNNs are able to extract patterns from massive volumes of complex data, and they become a natural candidate to tackle problems in the field of remote sensing. The use of deep embeddings based on CNNs is gaining increasing attention due to their ability to encode the images into relevant latent features used later in other classification or clustering methods \citep{ji2018, Kalinicheva20, Atar11}. However, the feature extraction obtained by the embedding is usually guided by a good compression of the information contained in the original image, and it is not directly related to a classification or clustering purpose. A way to enrich these kinds of models is to create semantically meaningful embeddings using tools such as the Tile2Vec algorithm \citep{Jean2019}. Specifically, this algorithm is learned in an unsupervised manner, aiming to generate an embedding where similar image patches have vector representations nearby while being distant from dissimilar patches. The term semantic refers to the meaningful comprehension and representation of complex spatial features and patterns within the imagery. A semantic deep embedding captures the inherent characteristics, relationships, and similarities among objects by encoding the relevant information into embedded vectors in such a way that the distances between them also hold significance.

In the current work, we propose a novel methodology that constitutes an advancement in several aspects not previously addressed in depth. Four main contributions are worth mentioning: i) we design a method to train the embedding considering time series instead of static images, therefore the final embedded vectors contain both spatial and temporal information, ii) we train the embedding from a grid of satellite images covering a large region, demonstrating the benefits of using embeddings to avoid several technical problems that arise in image fusion such as border effect, inconsistencies between sensors, or temporal shifts when using cloud-free images \citep{goyena2023unpaired}, iii) we go far beyond the classic pixel level in the creation of time series \citep{guyet2016,zhang2021} since we use multivariate time series (MTS) of embedded vectors that provide a much richer and more meaningful representation of the spatio-temoral dynamics of the land, iv) the methodology is fully unsupervised as relies on semantic embeddings and clustering of time series to identify areas, potentially large, that share both similar geographical characteristics and temporal evolution.


The procedure followed in this research can be summarized as follows. Firstly, based on the Tile2Vec approach, we adapt the training of the embedding to the SITS context. Secondly, the images are decomposed into square image patches, called tiles, and the embedding is used to obtain the vectorial representation of each tile. Then, the sequences of images are represented as a collection of MTS. Thirdly, we cluster the MTS using the $K$-means clustering method for time series. 
The final step of the methodology is to refine the embedding by resuming the training with information given by the previous clustering partition of MTS. Thus, the embedding is driven to assign a nearby representation in the embedded space to tiles sharing both similar semantics and temporal evolution. 

 The results derived from the proposed approach have been validated by an experienced geographer with deep knowledge of the region under study. This corroborates our claim that the proposed methodology yields semantically coherent divisions of land, enabling a comprehensive understanding of the terrain. The outcomes do not replicate image details as pixel-based partitions tend to do, but offers a higher abstraction level through coarse-grained clustering. 
This arrangement of the land is particularly distinguished by identifying structured patterns across wide areas that put together intricate evolving semantics such as river basins or mountain ranges, but also separate climatic zones or ecological systems.


The rest of the paper is organized as follows. 
Section \ref{sec:related} discusses relevant previous works. 
Section \ref{sec:tile2vec} summarizes the background of the Tile2Vec embedding. Section \ref{sec:method} develops the proposed methodology based on both the semantic embedding and the partitional clustering. Section \ref{sec:design} specifies the details of the experiments and the model parameters used for the case study. In Section  \ref{sec:results} we present and discuss the empirical results from the study. Finally, Section \ref{sec:conclu} draws conclusions and identifies possible future work.

\section{Related work} \label{sec:related}

Dealing with SITS data involves significant methodological and technical challenges. A crucial issue is a need for labeled data \citep{Storie18}, which are particularly difficult to obtain for satellite images. Moreover, even if the ground truth is available, the land cover class could change over time, especially when the time series is long \citep{guyet2016}. Although most of the work in SITS has been focused on supervised techniques for land cover classification, an increasing number of contributions are devoted to developing unsupervised methods. These proposals mainly work at the pixel level or rely on segmentation techniques. Thus, \cite{zhang2021} develop a procedure based on dynamic time wrapping (DTW) distance measures between time series of pixels and \cite{lampert19} carry out a constrained K-means clustering at pixel level that needs for a proportion of labeled data. Alternatively, \cite{khiali2019} characterize the dynamics of specific objects that evolve similarly, using the well-known segmentation algorithm Mean Shift \citep{Comaniciu02} to identify trackable objects within the SITS. The dynamics of the objects are represented using graphs, and the clustering algorithm is conducted on these graphs. For the experiments, they consider well-delimited areas up to a maximum extension of 95 Km$^2$, and the ground truth is provided by an expert. Similarly, \cite{Kalinicheva20} apply 3D convolutional autoencoders and a segmentation procedure also based on Mean Shift. They go a little beyond the pixel level by using patches of a maximum size of 9 pixels. The autoencoder is used to generate a vector representation that encodes spatio-temporal information from the whole time series of patches. Then, they use segmentation to find a unique segmentation map for the whole SITS. 

Relevant differences exist between the aforementioned contributions and the current research. Firstly, we work on a different scale, far beyond the pixel level. We focus on patches (a.k.a. tiles) of 100 pixels to capture broad and complex semantics belonging to potentially large regions. Secondly, we avoid the use of segmentation techniques for SITS which can be very context-specific, complex to use, and focused on fine-grained analysis. Instead, we let the partition naturally emerge from clustering. Thirdly, we put aside the study of DTW distance measures, since the benefits of using this kind of distance largely depend on the characteristics of the time series data sample. Since we consider regular time series with a low temporal resolution, there is no difference between using DTW or Euclidean distance.
Lastly, unlike \cite{Kalinicheva20} we consider the whole explicit time series of embedded vectors, which are generated by a semantic embedding specifically trained for this task. The use of autoencoders in \cite{Kalinicheva20} provides a good compression of the time series, but the distance between vectors has no special meaning. Therefore, the information contained in the time series and the relationships among them are obscured by the encoding that does not encourage vector discrimination. 
On the other hand, the use of segmentation leads to fine-grained modeling that, although highly suitable to classify well-defined elements such as roads, rivers, buildings, or crops, cannot be applied to analyze a region from a more global perspective i.e. from a higher level of abstraction that aims not to get lost into very specific details of the images. As previously introduced, we propose an alternative methodology to create partitions of the ground where large areas can potentially be grouped according to a more complex and general semantic. Our approach can reveal new global information about the land that is not possible to obtain with the aforementioned works. 
It uses multiple image sequences to capture spatio-temporal interconnections, such as those associated with ecological systems, climatic zones, forests, river basins, grasslands, or mountain ranges.  Thus, our method is able to get comprehensive terrain information that can be used for example as an intermediate step in fusion methods \citep{goyena2023unpaired} or it can be integrated into a preprocessing step before image fusion. Furthermore, prediction procedures for cloud filling, such as the IMA method introduced in \cite{militino2019interpolation}, where scalability may be problematic, can benefit from the use of intelligent clustering by reducing partition errors.

A crucial element of the current work is semantic embedding. 
The use of embeddings \citep{Taskin21} is gaining increasing attention in the field of satellite images, mainly due to the complexity of multi-spectral satellite image analyses. These methods are able to create embedded spaces where the images are encoded as vectors of bounded size. Here we are particularly interested in those that can create semantic embeddings, where not only the vectors are meaningful but also the distances between them represent the amount of dissimilarity between the images. The embedding built for the current work is based on the Tile2Vec model   \citep{Jean2019}.
It uses simple arithmetic operations within the space that conserve semantic properties. This provides a solid basis for analyzing the final clustering partition obtained by the proposed methodology from a semantic perspective.
The main contribution of \cite{Jean2019} is how the model is learned. They use a triplet loss function where triplets of tiles are generated according to a spatial neighborhood, providing a kind of weak supervision. Thus, similar tiles are mapped close to each other, and different tiles are mapped as far apart as possible.
In the current research, we use the CNN of Tile2Vec and its loss function, but we generate the triplets in a specific way to bring the model into the context of SITS and ultimately, to assist in identifying areas with both similar semantic properties and similar temporal evolution. To the best of our knowledge, semantic embeddings have not been specifically trained and analyzed within the context of SITS clustering.
In \cite{Wozniak2021} the authors present an approach analogous to Tile2Vec but using hexagonal instead of squared tiles. Alternatively, \cite{Jung2022} use a SimCLR approach, an encoder network trained to maximize agreement by using contrastive loss \citep{Chen2020}, and modify this model to include $k$-neighbor tiles, but no distant tiles are considered. However, we use the Tile2Vec approach since it has demonstrated that within the resultant embedded space, basic operations between vectors conserve the semantic meaning. This is a very useful property for our posterior clustering analysis.

\section{Tile2Vec in a nutshell} \label{sec:tile2vec}
Tile2Vec \citep{Jean2019} is a semantic embedding method 
that departs from using labeled images and instead derives image semantics from Tobler's first law of geography \citep{tobler1979}: \textit{everything is related to everything else, but near things are more related than distant things}. As atomic units, Tile2Vec considers tiles, $x$, of fixed dimension as square image patches taken from a satellite image $X$. Following Tobler's law, the algorithm assumes that, on average, closer tiles are more similar than distant tiles, and therefore, their embedded representation has to be close. The learning process is expected to build not only an embedded space where vectors of similar images are closer to each other than vectors of dissimilar images but also to capture the corresponding degree of similarity.

Tile2Vec is learned from a training set of triplets of tiles $(x_a,x_b,x_c)$, where $x_a$ denotes the anchor tile, $x_b$ the neighbor tile, and $x_c$ the distant tile. 
Given a spatial neighborhood of $r$ pixels, the neighboring tile's center must be within this range from the anchor tile. The distant tile is randomly selected from outside this area.
The embedding function, f, is given by a ResNet-18~\citep{He16} CNN with parameters $\theta$, a modified input to handle multi-spectral tiles, and without the final classification layer.
Thus, $f$ maps a tile $x$ to a $d$-dimensional vector $z \in \mathds{R}^d$. The loss for a triplet $(x_a,x_b,x_c)$ is given by:
\begin{eqnarray}
\label{eq:loss}
L(x_a,x_b,x_c|\theta) = [||f(x_a|\theta) - f(x_b|\theta)||_2 - \nonumber \\
||f(x_a|\theta) - f(x_c|\theta)||_2 + \delta]_+
\end{eqnarray}
\noindent where $[\cdot]_+$ is the positive part of the argument. $\delta \geq 0$ is the margin, which bounds the difference between the two distances of the triplet loss function. Otherwise, the model could increase the distance between the anchor and the distant embedding without restriction to minimize the loss.
The objective function is the sum of loss for the whole training set of $N$ triplets $D=\{(x^{(1)}_a,x^{(1)}_b,x^{(1)}_c), \ldots, (x^{(N)}_a,x^{(N)}_b,x^{(N)}_c)\}$:
\begin{eqnarray}
\label{eq:objectiveFunction}
     \nonumber
     \min_{\bm{\theta}} \sum_{i=1}^N \left[ L(x_a^{(i)},x_b^{(i)},x_c^{(i)}|\theta) + \right. \\
     \lambda \left. \left(||f(x_a^{(i)}|\theta)||_2 + ||f(x_b^{(i)}|\theta)||_2 +  ||f(x_c^{(i)}|\theta)||_2 \right) \right] 
\end{eqnarray}
where $\lambda$ is a regularization parameter that bounds the resultant vector magnitudes. 
In summary, the learning algorithm finds the embedding function $f$ that minimizes the Euclidean distance between an anchor and its neighbor tile while maximizing the Euclidean distance between the anchor and the distant tile. For further details see~\cite{Jean2019}.

Tile2Vec-based models naturally emerge as suitable candidates to build \textit{multidimensional time series embedding for sequences of tiles} on which it makes sense to perform and interpret partitional clustering.  Nonetheless, it's worth highlighting that before clustering, a crucial step involves adapting the training approach of Tile2Vec to encompass both geographic and temporal information (see Sections \ref{sec:geographical_embedding}  and \ref{sec:clustering_embedding}).

\section{Methodology}  \label{sec:method}
 
The spatio-temporal clustering of the satellite images is carried out on the embedding of sequences of tiles. This codification is scalable and interpretable and constitutes a fundamental element of the methodology.
We assume that each satellite image can be decomposed into small tiles that contain relevant pieces of geographic information when considered in isolation. Intuitively, the size of a tile should be the minimum to capture patterns such as those shown by rivers, mountains, hills, crops, or pastures. 

Let $X=\{x_1,...,x_m\}$ be an image of a region of interest that we assume is decomposed into a grid of square tiles, $x_i$, for $i=1...m$. Note that $X$ can also be composed of a set of satellite images covering the region of interest. Let $(X^1,...,X^T)$ be a temporal sequence of images of the same region, where $X^t$ is the image of the region at time $t$, for $t=1,...,T$. From these images, we get sequences of tiles  $\bd{X}=\{\bd{x}_1,...,\bd{x}_m\}$, where $\bd{x}_i=(x_i^1,...,x_i^T)$ corresponds to the $i$-th sequence of tiles.

Using the semantic embedding, we represent sequences of tiles as MTSs of embedded vectors. Thus, we propose the following two-step procedure to perform the clustering:

\begin{enumerate}
    \item \textit{Geographic-based land partitioning:}
    \begin{enumerate}
        \item[1.1] Learn the geographic-based embedding, $f^g$, from  $(X^1,...,X^T)$ (Section \ref{sec:geographical_embedding}).
        \item[1.2] Embed the set of sequences of tiles $\bd{X}=\{\bd{x}_1,...,\bd{x}_m\}$ with $f^g$ and cluster the resultant MTSs (Section \ref{sec:clustering}).
    \end{enumerate}
    \item \textit{Clustering-based land partitioning:}
    \begin{enumerate}
        \item[2.1] Learn the clustering-based embedding, $f^c$ (Section \ref{sec:clustering_embedding}).
        \item[2.2] Embed the set of sequences of tiles $\bd{X}=\{\bd{x}_1,...,\bd{x}_m\}$ with $f^c$ and cluster the resultant MTSs.
    \end{enumerate}
\end{enumerate}

The procedure starts in 1.1 by learning the embedding $f^g$ using a training set of triplets taken from the available sequences of images, $(X^1,...,X^T)$. Each triplet in the dataset belongs to the same time. Neighbor and distant tiles are defined according to a spatial distance. Using the embedding function, $f^g$, we represent each sequence of tiles, $\bd{x}_i$, as an MTS, $\bd{z}^g_i=(f^g(x_i^1),...,f^g(x_i^T))$, for $\bd{x}_i \in \bd{X}$. 
Then, in 1.2, we obtain a clustering partition of $\bd{Z}^g=\{\bd{z}_1^g,...,\bd{z}_m^g\}$, $\s{P}^g=\{P_1^g,...,P_K^g\}$ with $P_k^g \subset \bd{Z}^g$ for $k=1,...,K$. In 2.1, we learn a second embedding $f^c$ from triplets whose tiles are sampled according to the neighborhood given by $\s{P}^g$ so that the neighbor tile belongs to the same cluster as the anchor tile and the distant tile is chosen from a different cluster.
The clustering-based embedding $f^c$ constitutes a refinement of the geographic-based embedding $f^g$. Since the clustering, $\s{P}^g$, captures spatio-temporal patterns, the triplets sampled from $\s{P}^g$ include additional spatial and temporal information. Thus, the neighbor tiles tend to be semantically similar to the anchor tiles and additionally, they belong to regions that change similarly over time. On the other hand, distant and anchor tiles tend to be semantically different and/or belong to regions that change differently over time. This step aims at reinforcing the target of the clustering, which is, in essence, to group areas with a similar evolving semantic.  
Finally, in 2.1, the function $f^c$ is used to obtain a new embedded representation for each sequence of tiles $\bd{x}_i$ as $\bd{z}^c_i=(f^c(x_i^1),...,f^c(x_i^T))$, for $i=1,...,m$. Thus, we can obtain the final clustering partition. In the rest of this section, we provide a detailed explanation of our proposal.

\subsection{Geographic-based embedding}\label{sec:geographical_embedding}

We use the objective function defined in \ref{eq:objectiveFunction} to obtain the semantic embedding. Nevertheless, we must develop an appropriate method to generate the training set of triplets, $D^g$, in the context of time series. Since we want to capture the semantics of a region as a whole for any given time, we observe the following restriction: the tiles within a triplet must belong to images from the same time $t$. Due to this temporal constraint, intuitively, our embedding is only focused on the semantics of the spatial component of every image within the sequence. The temporal information is later introduced by using multidimensional time series to encode the sequences of tiles.

When large geographical areas are analyzed, $X$ may be a composition of several satellite images needed to cover the whole region of interest. As our method deals with the semantics of the tiles, these individual satellite images do not require any regularization or pre-processing step to be used. This is a noteworthy advantage of the proposed method since the composition of satellite images usually produces border effects that may alter the posterior analysis.


The training set $D^g$ consists of $N$ triplets. To obtain the neighbor and distant tiles in a triplet, we consider a neighborhood around $x_a$ (the anchor tile). This neighborhood is defined by a circle with a radius of $r$ pixels centered at $x_a$. The value of $r$ is typically set to be equivalent to the tile size to allow for partial overlap or close geographical proximity between anchor and neighbor tiles. Thus, anchor and neighbor tiles are very likely to have similar semantics. Consequently, each triplet is obtained using the following procedure:
\begin{itemize}
    \item Select $t$ at random from $\{1,\ldots,T\}$.
    \item Select an anchor tile, $x_a^t$, at random from $X^t$.
    \item Select a neighbor tile, $x_b^t$, at random within the neighborhood of $x_a^t$ in $X^t$.
    \item Select at random from $X^t$ a distant tile, $x_c^t$, located outside the neighborhood of $x_a^t$.
\end{itemize}

Once $D^g$ is created, we learn the geographic-based embedding function $f^g$ from $D^g$.

\subsection{Time series Clustering}\label{sec:clustering}

Given a set of embedded sequences of tiles, $\bd{Z}=\{\bd{z}_1,....,\bd{z}_m\}$, we aim to identify $K$ groups by using partitional clustering techniques. In particular, we propose to solve the $K$-means problem for $\bd{Z}$, where $K$ determines the number of clusters. The $K$-means problem consists of finding a partition $\s{P}=\{P_1,...,P_K\}$ (with non-empty clusters) that minimizes the error:
\begin{equation}
E(\s{P})=\sum_{k=1}^K \sum_{\bd{z} \in P_k} d(\bd{z}_k,\bd{c}_k)^2,\label{eq:kmeanserror}
\end{equation} 
where $d(\bd{z},\bd{z}')=\sum_{t=1}^T ||z_t - z_t'||_2$ is the Euclidean distance between the MTSs $\bd{z}$ and $\bd{z}'$, and $\bd{c}_k=\frac{1}{|P_k|}\sum_{\bd{z} \in P_k} \bd{z}$ is the centroid of the cluster $P_k$ which corresponds to the average of the MTS within this cluster. 

The use of the $K$-means algorithm \citep{lloyd1982least} is motivated by several reasons. It is an iterative procedure that generates a sequence of clustering partitions with a monotone decreasing error function (Equation~\ref{eq:kmeanserror}) until convergence is reached. The algorithm is linear in the number of considered data points, $m$, and therefore, the proposed methodology can be applied to a large collection of MTS. Since $K$-means utilizes centroids, they can be used as reference points to obtain the representative sequence of tiles for each cluster.
\cite{Jean2019} show that the interpolation of two hand-picked tiles embedded with Tile2Vec allows for covering the full spectrum of intermediate patterns. We exploit this property in a fully automatic process by using the centroids provided by $K$-means and the interpolation between them. This is feasible because the $K$-means algorithm favors the formation of spherical or convex clusters, where the convex hulls of individual clusters are not intertwined. The convex hull refers to the smallest convex shape that encompasses all the points within a cluster.
Convex clusters are particularly interesting from the embedding perspective because the convex combination of any subset of points from a cluster $P_k \in \s{P}$ belongs to the convex hull of $P_k$. In other words, both the centroid of the cluster $P_k$ and any other point obtained by linear interpolation between two points within $P_k$ belong to the class of points given by the cluster $P_k$. In the context of distinct clusters, the linear interpolation between two points is found to intersect the border in the embedding space precisely once. In essence, this implies a smooth transition from one cluster to another.
In simpler terms, as we move from one cluster to the adjacent one, the points in that path gradually change from the characteristics of the first cluster to those of the second cluster, ensuring a seamless and continuous progression. Additionally, Tile2Vec loss-function and $K$-means error are related. These make $K$-means the preferred option among other clustering algorithms.

\subsection{Clustering-based embedding} \label{sec:clustering_embedding}
The last step consists of refining the geographical-based embedding $f^g$ by using information obtained from the clustering partition $\s{P}^g$ and conducting the final clustering partition of sequences of tiles using the new embedding $f^c$. 
For this purpose, we generate a new training set of triplets $D^c$ using a neighborhood based on $\s{P}^g$. With an abuse of notation, in this section, we consider that $\s{P}^g$ corresponds to the partition of the sequences of tiles from the original images, $\{\bd{x}_1,...,\bd{x}_m\}$, associated to the geographic-based embedding $\bd{Z}^g$. The triplets conforming $D^c$, $(x_a^t,x_b^t,x_c^t)$, again satisfy the aforementioned temporal constraint. The training set $D^c$ consists of $M$ triplets as follows:
\begin{itemize}
    \item Select a cluster index $k$ at random from $\{1,...,K\}$ with probability proportional to the size of the cluster $|P_k^g|$.
    \item Select an anchor sequence $\bd{x}_a$ uniformly at random from the cluster $P_k^g$.
    \item Select a neighbor sequence $\bd{x}_b \neq \bd{x}_a$ uniformly at random from the cluster $P_k^g$.
    \item Select at random a cluster $P_j^g$ (with $j\neq k$) with a probability proportional to $|P_j^g|$ and $d(\bd{c}_k,\bd{c}_j)$.
    \item Select a distant sequence $\bd{x}_c^t$ uniformly at random from $P_j^g$.
    \item Construct the triplet $(x_a^t,x_b^t,x_c^t)$ by selecting $t$ uniformly at random from $\{1,...,T\}$.
\end{itemize}
Then, we obtain the clustering-based embedding $f^c$ by tuning $f^g$ parameters with the new training $D^c$.
The obtained clustering-based embedding, $f^c$, is still a mapping from the tile space $\s{X}$ into $\mathds{R}^d$. The clustering-based embedding aims to reduce the average intra-cluster dissimilarity of $\s{P}^c$ with respect to the geographical-based embedding while increasing the average inter-cluster dissimilarity.

Finally, the sequences of tiles are re-clustered using $K$-means in $\bd{Z}^c=\{\bd{z}_1^c,...,\bd{z}_m^c\}$, providing the final clustering partition (see Section \ref{sec:clustering}). This final partition is considered a refinement of the clustering partition obtained from $\bd{Z}^g$.

\section{Design of the experiments} \label{sec:design}

In this section, we illustrate our proposal by using a sequence of Sentinel-2 images located around the province of Navarre (northern Spain) that contains relevant areas, such as the Pyrenees or the Ebro river basin.  We chose this region due to its environmental variability within just 220 km$^2$. In this region, we can find the Cantabrian valleys in the north, which have a temperate and humid climate, with abundant cloudiness and precipitation. On the contrary, in the southern part, the continental Mediterranean climate appears, arid and dry in the Ribera Navarra area, which takes on desert-like features in the Bardenas. Next, we provide the details of the learning parameters, the satellite imagery dataset, and the tools to analyze the results.

\subsection{Image dataset and training parameters}\label{sec:training}
We use Sentinel-2 RGB-bands to create images of size $10980\times 10980$ pixels with a spatial resolution of $10$ meters per pixel. These three bands, along with the near-infrared, are the only bands provided by Sentinel-2 with this resolution. Since this work stresses the use of MTSs and their interpretation, we decided to deal with images of bounded complexity in terms of band composition. 

\begin{figure}[!t]
	\centering
    \includegraphics[width=\linewidth]{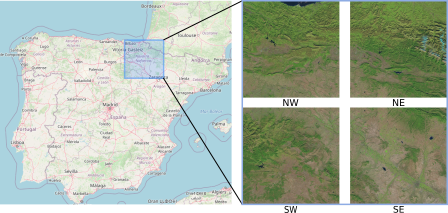}
	\caption{Region chosen for experiments in northern Spain, covered by 4 Sentinel-2 images}
	\label{fig:region}
\end{figure}

The area selected to illustrate the proposed methodology is depicted in Figure \ref{fig:region}. The training of the first embedding is carried out with sequences of images of the four regions indicated on the right of this figure. The whole selected area contains a wide variety of land types such as high and low mountains, crops, pastures, or rivers, and it exhibits different characteristics throughout the year, such as snow-covered areas or harvested fields. Thus, extracting and synthesizing the most relevant information of such a complex and large environment may demonstrate that the methodology presented in this paper works in practice. The proposed methodology provides a general framework and can be applied to any other places, resolutions, and bands.
In this particular case, we get images of each season of the year for five years (2017-2021). Therefore, we use a total of $4 \mbox{ (regions) } \times 5 \mbox{ (years) }\times 4 \mbox{ (seasons) } = 80$ Sentinel-2 images to train the first embedding $f^g$.

We first train the geographic-based embedding, $f^g$, following the procedure proposed in Section \ref{sec:geographical_embedding}. A total of $N=100000$ triplets sampled from the Sentinel-2 images, $5000$ triplets from each timestamp, are used. A training set with $100k$ triplets has shown to be large enough to obtain a rich semantic representation of the image semantics \citep{Jean2019}. The size of the tiles is $100\times 100$ pixels (covering 1 km$^2$), and the geographical neighborhood is given by a ball of radius $r=100$ pixels whose center is the center of the anchor tile. 

Given the Sentinel-2 images, tiles of $100\times 100$ pixels allow capturing relevant spatial characteristics of the region. The training process is iterated $50$ epochs as the convergence of the loss function is observed. The batch size is set to $50$, the margin is $\delta= 50$, and the regularization parameter is $\lambda =0.01$. Since we are interested in the relative distances among vectors, the values of $\delta$ and $\lambda$ are not crucial elements in this case, so they are set as it is done in \cite{Jean2019}.
We set the last layer of the network, i.e. the dimensions of the embedding space, to 512 features to have a trade-off between a rich representation of the tiles and an affordable dimension. For the clustering-based embeddings $f^c$, we re-train the model as described in Section \ref{sec:clustering_embedding}, with $M=20000$ triplets. The number of triplets has been reduced in this case to properly combine both models. 
The neighborhood is given by a partitional clustering of size $K=5$ since this number of clusters is suitable for human interpretation in the context of the current research, but it could be fitted according to the problem at hand. The training process in this case is iterated $25$ epochs, which is enough for the model to converge. Note that the computational cost to train the embeddings mirrors that of the Tile2Vec algorithm \citep{Jean2019}.

\subsection{Tools of analysis} \label{sec:tools}

In this section, we introduce the tools used to analyze, validate, and evaluate the results obtained when using the proposed methodology. These tools include visualization techniques for portraying clustering, a comparative study between embedding-based and pixel-based clustering, and an exploration of semantic information captured by the clustering.

\paragraph{Geographic representation}

We plot the original satellite images as color-maps according to the clustering of MTSs obtained with $K$-means, $\s{P}=\{P_1,..., P_K\}$. The color of each cluster is given by its corresponding centroid, $\bd{c}_k$, as follows. The colors are automatically generated using principal component analysis (PCA) \citep{Jolliffe2002}. The PCA is fitted with the entire set of MTSs, $\bd{Z}$, and then, we extract the first three components of each centroid. In our specific case, the first three principal components are utilized as RGB values for visualization purposes. Thus, each cluster $P_k$ is represented by its centroid, $\bd{c}_k$, and its color is given by the first three PCA components of $\bd{c}_k$.
As the centroid captures the overall semantics of the sequences of tiles within the cluster, if two clusters have similar colors (in the RGB space) it means that they are semantically similar, and vice versa. This representation facilitates the interpretation of the clustering partition and some intuition about the general spatio-temporal pattern of the region and, in particular, about the possible number of clusters behind the images. 

\paragraph{Embedded representation}
As a complement to the previous geographical representation, we show the clustering partition through a two-dimensional projection of the embedded space. 
Specifically, the original embedded space is projected down to two dimensions by using Multidimensional Scaling (MDS) \citep{kruskal1964MDS}.
The distance used is the Euclidean distance between MTS, the same distance as in Equation \ref{eq:kmeanserror}. Then, each point in the two-dimensional space corresponds to an MTS, and it is depicted with the color of the cluster to which it belongs. 
MDS is a deterministic algorithm that seeks a low-dimensional representation of the data that preserves the relative distances of the high-dimensional embedded space. The two-dimensional projections are used to compare the geographical-based and the clustering-based embeddings.

\paragraph{Pixel-based clustering}
Working at the pixel level has been widely used in the research literature on satellite imagery. Since, to the extent of our knowledge, there is no other method devoted to a semantic large-scale analysis as the method we propose here, we believe it is worth comparing the clustering presented in the current work with pixel-based clustering. Nonetheless, examining each individual pixel's time series is impractical and cost-prohibitive for large-scale analysis. Therefore, we compute the average RGB components of pixel tiles. Specifically, we consider tiles consisting of 8 pixels each, as it produces a dataset for clustering of the same size as the one obtained with the MTSs used in our proposal. We also use the average RGB components of tiles of $100\times100$ with the specific purpose of comparison with the embedding-based clustering by two standard clustering quality measures: Silhouette and Calinski-Harabasz scores  \citep{ROUSSEEUW198753, calinski74}. To ensure a fair quantitative comparison between the embedding-based and pixel-based clusterings, these scores are calculated using distances in the two-dimensional space of the original images, specifically based on the 2D coordinates of the tiles. This approach allows for comparison within the same geographic space for both clustering methods. Note that the silhouette score ranges from $-1$ to $1$, where a high value indicates that the object is well matched to its own cluster and poorly matched to neighboring clusters.



\paragraph{Representatives, interpolations, and semantic tree}

We introduce three essential tools designed to validate the effectiveness of our unsupervised approach in capturing accurate, valuable, and structured spatio-temporal information. These experiments rely on the significance of distances within the embedding.
\begin{itemize}
    \item \textit{Cluster representative}: the sequence of tiles, $\bd{x}_k$, that is closest to the centroid $\bd{c}_k$ within the cluster. This central point encapsulates the essence of the cluster, providing a symbolic representation of the data within.
    \item \textit{Interpolation}: the intermediate MTS, $\bd{z}_w= w\cdot \bd{c}_k+ (1-w)\cdot \bd{c}_{k'}$, for $w \in [0,1]$, between pairs of centroids $\bd{c}_k$ and $\bd{c}_{k'}$. We show the representative sequence of tiles, $\bd{x}_w$ (the closest sequence to $\bd{z}_w$). The interpolations aim to bridge the gap between distinct clusters by generating transitional representations between their centroids. They offer clear and coherent insights into the connections among clusters within a complex and intricate geographic system. 
    \item \textit{Semantic tree}: the minimum spanning tree \citep{prim1957shortest} of a complete undirected graph calculated from the distance matrix among cluster centroids. This tree is essential for linking clusters and guiding the aforementioned interpolation experiment. Through this approach, we can reveal the primary connections among clusters on a global scale, offering a structured and interpretable visualization of the entire region in terms of spatio-temporal relationships. 
\end{itemize}
This set of experiments are designed to be interpretable and easily reviewed by experts.
 
\section{Results and analyses} \label{sec:results}

This section presents the results of the aforementioned experiments. Firstly, we study the clustering partitions generated with the geographic-based embedding, $f^g$, and compare them with pixel-based clusterings. Secondly, we study the impact that the clustering-based embedding, $f^c$, has on the underlying structure of the clustering partition. Lastly, we inspect the semantics captured by the cluster centroids and the interpolations between them. For the sake of clarity and to conserve space, we only show the most relevant findings. All the results have been validated by a field expert with in-depth knowledge of the entire area. Official geographic information sources have also been consulted \citep{idena, signa, bio}.

\subsection{Geographic and embedded representations}

\begin{figure*}[!t]
\centering
\subfloat[NE region]{\includegraphics[width=0.245\linewidth]{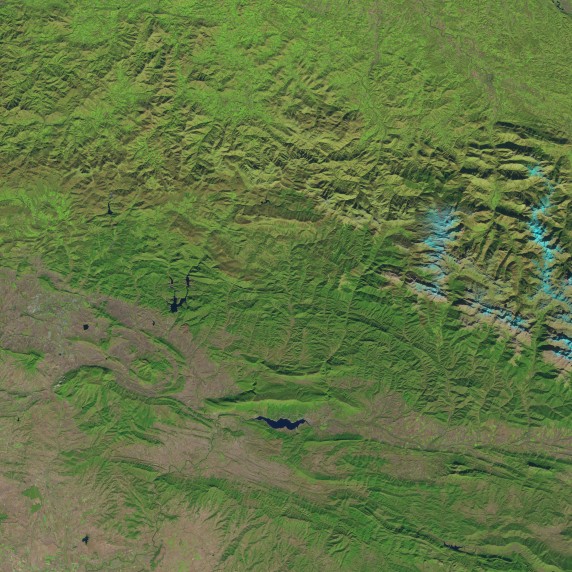}}
\subfloat[$K=3$]{\includegraphics[width=0.25\linewidth]{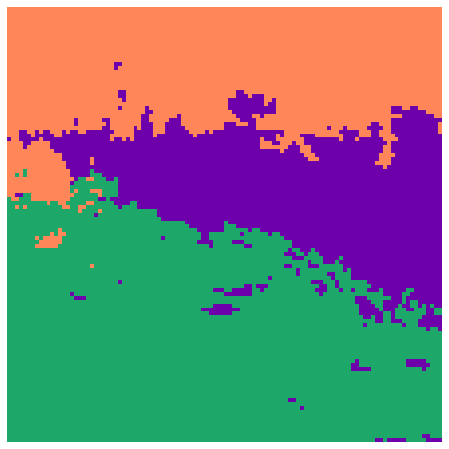}}
\subfloat[$K=4$]{\includegraphics[width=0.25\linewidth]{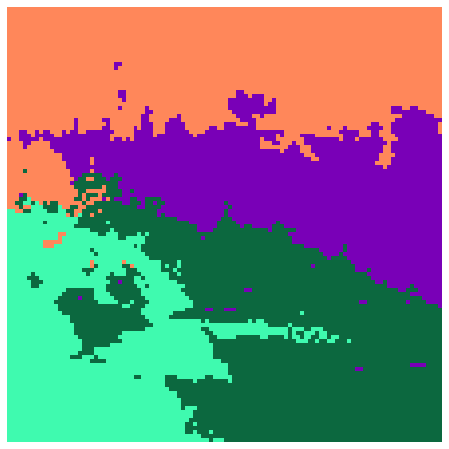}}
\subfloat[$K=5$]{\includegraphics[width=0.25\linewidth]{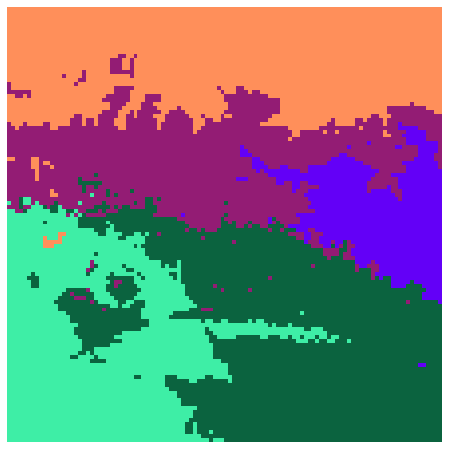}}\\
\subfloat[Collection of MTS]{\includegraphics[width=0.25\linewidth]{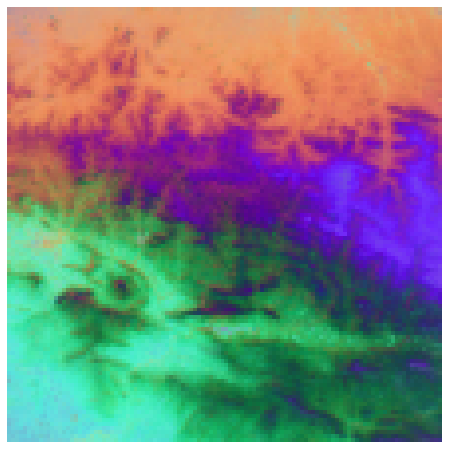}}
\subfloat[$K=3$]{\includegraphics[width=0.25\linewidth]{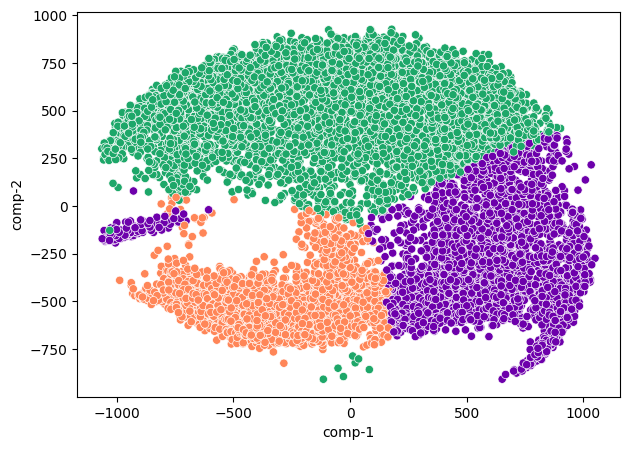}}
\subfloat[$K=4$]{\includegraphics[width=0.25\linewidth]{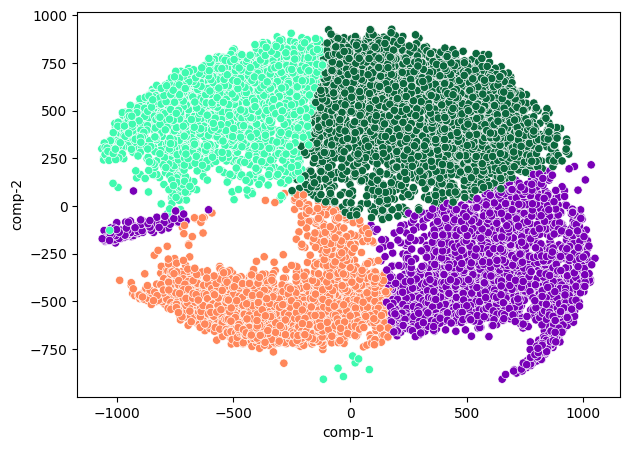}}
\subfloat[$K=5$]{\includegraphics[width=0.25\linewidth]{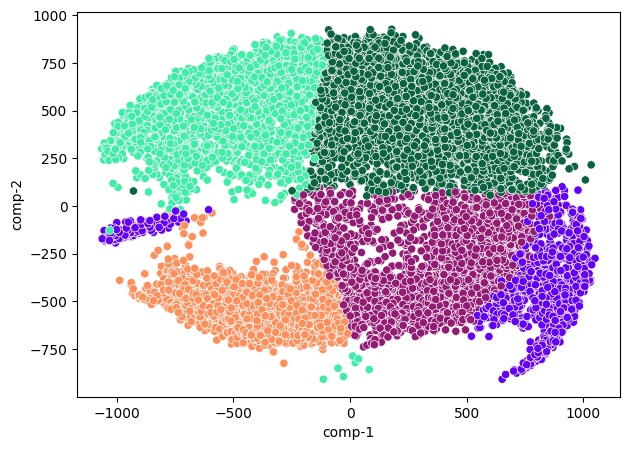}}
\caption{Geographical representations and 2D projections of the clustering $\s{P}^g$ with different number of clusters.}
\label{fig:clustering_geo}
\end{figure*}

In this section, we explore the geographic-based embedding, $f^g$, and the partitions $\s{P}^g$ generated from this embedding. The first results are shown in Figure \ref{fig:clustering_geo}. In this case, we focus on the NE region, given that identical behavior patterns are displayed across the remaining regions. Figure \ref{fig:clustering_geo} (a) shows one of the $20$ images of the temporal sequence of this region. The mountains in the middle of this image correspond to the Pyrenees, where we can see snowy mountains on the east side.

Figure \ref{fig:clustering_geo} (e) shows the land in the light of the embedding $f^g$, before any clustering analysis. The color map is generated automatically by assigning a different color to each MTSs according to PCA, as explained before. This chart condenses a great deal of information regarding the semantics of the region, its evolution over time, and the relationships among different zones. From an overall perspective, three big areas can be clearly distinguished: 
the northern part of the Pyrenees (shades of oranges), the southern part (shades of greens), and the Pyrenees themselves (shades of purples).
This division is concisely captured by the clustering with $K=3$ shown in Figure \ref{fig:clustering_geo} (b), whose corresponding MDS 2D projection is shown in Figure \ref{fig:clustering_geo} (f). These three areas are also associated with three different climatic zones: oceanic climate (orange), continental-Mediterranean climate (green), and alpine climate (purple).

Figure \ref{fig:clustering_geo} also shows the results of the clusterings with $K=4$ and $K=5$ in the charts (c, d, g, h).  We can observe that a hierarchical pattern naturally emerges. The big areas are mostly kept intact, and they are subdivided as the number of clusters increases, revealing new large and well-defined zones. Thus, the green cluster of Figure \ref{fig:clustering_geo} (b), which is related to a continental climate on the Mediterranean slope, is divided into two clusters in Figure \ref{fig:clustering_geo} (c) with $K=4$. In this case, the $\lgreen$ cluster is mostly associated with cultivated areas of Mediterranean rainfed agriculture, which are characterized by a wide variety of crops such as cereals, vineyards, or olives, although pastures can also be found. The $\dgreen$ cluster primarily encompasses non-cultivated regions, including pastures, forests, mountain vegetation, and areas dedicated to mountain agriculture.
In Figure \ref{fig:clustering_geo} (d), with $K=5$, it is important to note that the highest peaks of the Pyrenees, belonging to the axial zone ($\vblue$), are identified within the wider mountainous Pyrenean area ($\wine$). In general, we can observe a very structured pattern, compactly grouping large meaningful areas. The method is able to abstract from the small details of the images and, as the number of clusters increases, find out more specific semantics within the clusters.
The 2D projections indicate that the clusters are also well-defined in the embedded space and provide complementary information regarding the morphology of the clusters and the neighboring relationship between them.

\begin{figure*}[!t]
\centering
\subfloat[NE region]{\includegraphics[width=0.248\linewidth]{figures/NE.jpg}}
\subfloat[Pixel-based $8\times8$]{\includegraphics[width=0.25\linewidth]{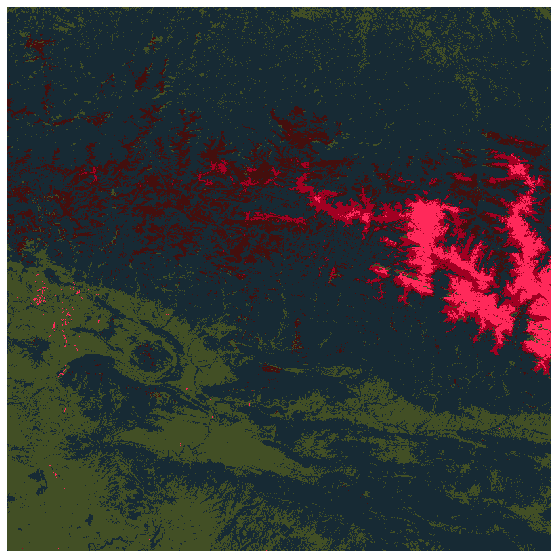}}
\subfloat[Pixel-based $100\times100$]{\includegraphics[width=0.25\linewidth]{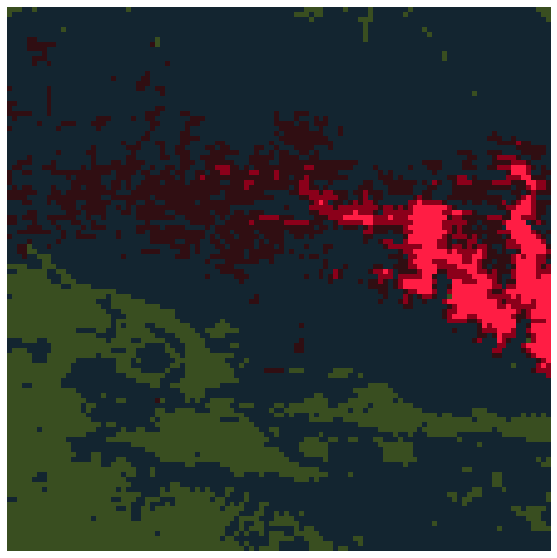}}
\subfloat[Embedding-based]{\includegraphics[width=0.25\linewidth]{figures/pca_k5_f1.png}}
\caption{Comparison between pixel-based and embedding-based clusterings for NE region. (a) Representative satellite image of the sequence. (b) Pixel-based clustering with tiles of size $8\times8$ pixels. (c) Pixel-based clustering with tiles of size $100\times100$ pixels. (d) Embedding-based clustering $\s{P}^g$.}
\label{fig:clustering_pixel1}
\end{figure*}

\begin{figure*}[!t]
\centering
\subfloat[NW region]{\includegraphics[width=0.27\linewidth]{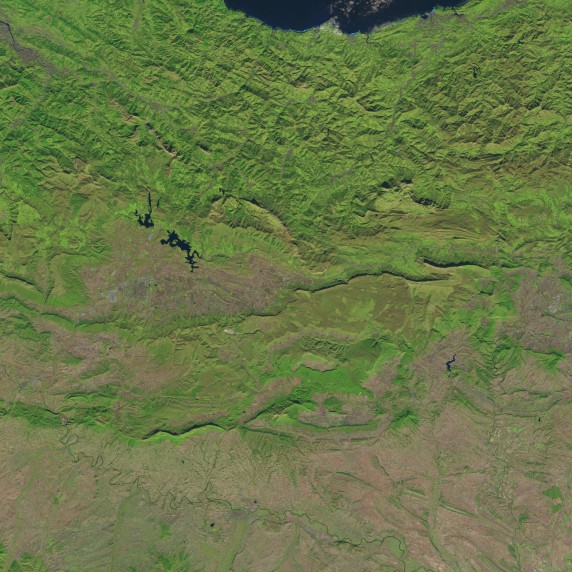}}
\hspace{0.5cm}
\subfloat[NW pixel-based]{\includegraphics[width=0.27\linewidth]{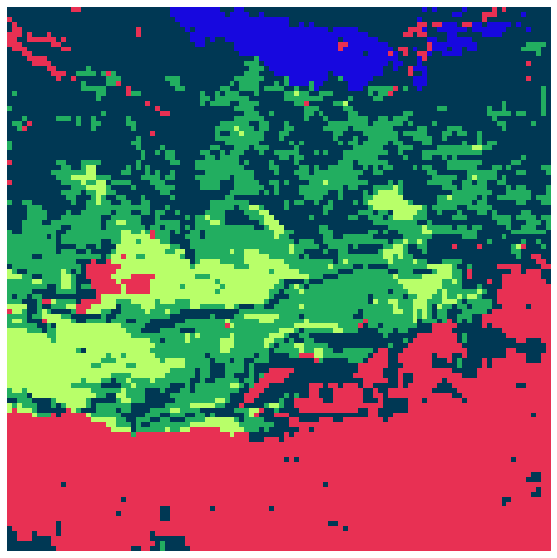}}
\hspace{0.5cm}
\subfloat[NW embedding-based]{\includegraphics[width=0.27\linewidth]{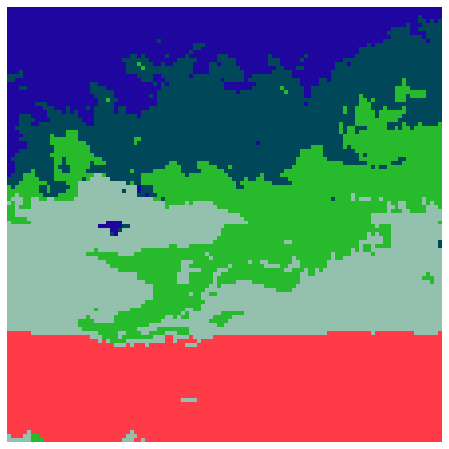}}\\
\subfloat[SW region]{\includegraphics[width=0.27\linewidth]{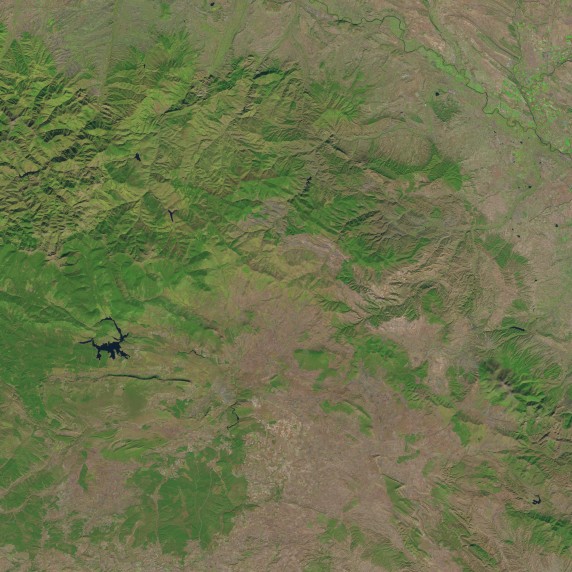}}
\hspace{0.5cm}
\subfloat[SW pixel-based]{\includegraphics[width=0.27\linewidth]{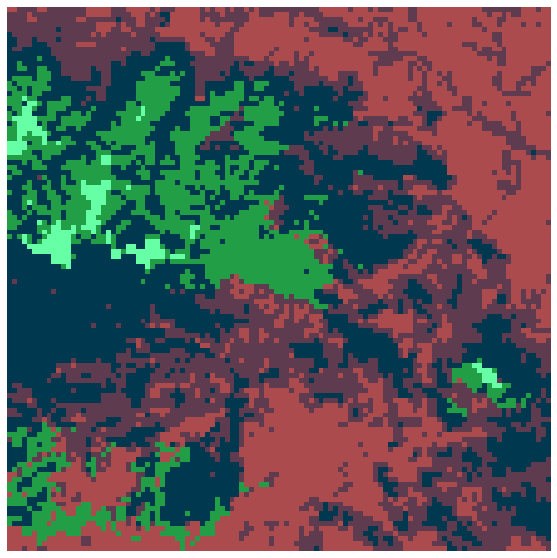}}
\hspace{0.5cm}
\subfloat[SW embedding-based]{\includegraphics[width=0.27\linewidth]{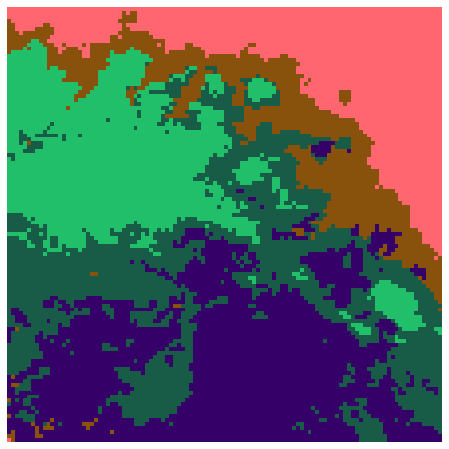}}\\
\subfloat[SE region]{\includegraphics[width=0.27\linewidth]{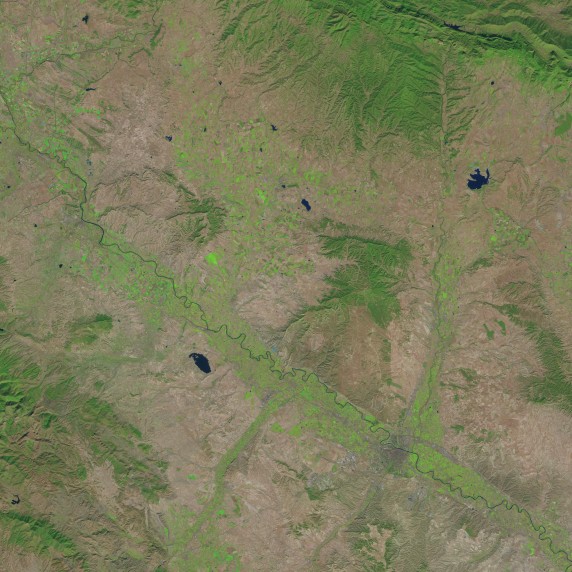}}
\hspace{0.5cm}
\subfloat[SE pixel-based]{\includegraphics[width=0.27\linewidth]{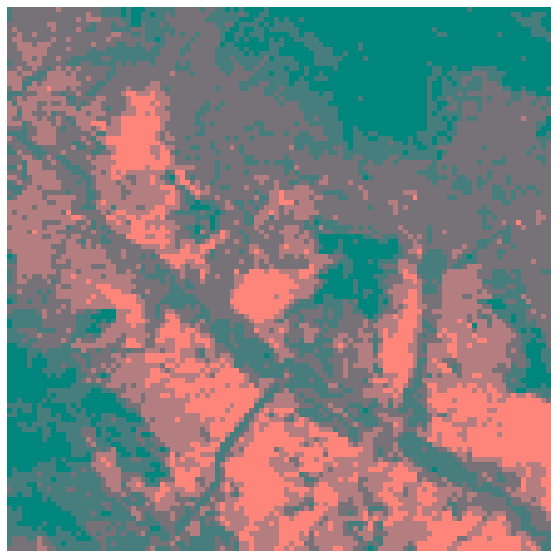}}
\hspace{0.5cm}
\subfloat[SE embedding-based]{\includegraphics[width=0.27\linewidth]{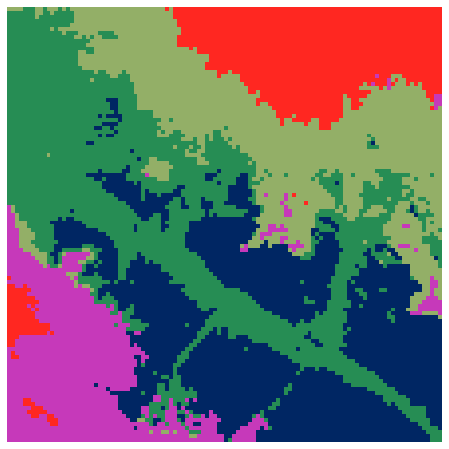}}
\caption{Comparison between pixel-based and embedding-based clusterings. The first column shows satellite images representatives of the sequence, the second column shows the pixel-based clusterings with tiles of size $100\times100$ pixels and the last column shows the embedding-based clusterings $\s{P}^g$.}
\label{fig:clustering_pixel2}
\end{figure*}

As we have seen so far, the land partitions given by the clustering of MTS are characterized by showing clearly separated and easily identifiable areas. The clusters obtained cover compact and large areas in the geography to which a field expert can easily assign semantic properties from a high level of abstraction. To validate this pattern of behavior, we compare the clusterings based on embedded sequences of tiles with clusterings based on pixels for the four available regions, choosing $K=5$ for this experiment. Figures \ref{fig:clustering_pixel1} and \ref{fig:clustering_pixel2} show a visual comparison, while Table \ref{tab:clustering_pixel} provides quantitative results. We firstly show the NE region in Figure \ref{fig:clustering_pixel1}, mainly to analyze the two versions of the pixel-based clusterings mentioned before, and compare them with the clustering partition, $\s{P}^g$, given by the embedding. We confirm that the pattern in which the terrain is arranged for the two pixel-based clusterings is similar, as can be seen in Figures \ref{fig:clustering_pixel1}(b) and (c). This result is analogous for all regions (not shown here) and allows us to verify that the pixel-based clustering with tiles of $100\times 100$ is suitable for comparison.

\begin{table}
\caption{Numerical comparison between pixel-based and embedding-based clusterings.}\label{tab:clustering_pixel}%
  \begin{tabular}{@{}lcccc@{}}
    \toprule
    \multirow{2}{*}{} &
      \multicolumn{2}{c}{Embedding-based} &
      \multicolumn{2}{c}{Pixel-based} \\
      \cmidrule(lr){2-3} \cmidrule(lr){4-5}
    Region & Silhouette & C-H & Silhouette & C-H  \\
    \midrule
    NW & 0.11 & 2683  & -0.05 & 1458 \\
    NE & 0.14 & 4463  & -0.14 & 750 \\
    SW & 0.09 & 2504  & -0.11 & 348 \\
    SE & 0.10 & 2320  & -0.06 & 314 \\
    \botrule
  \end{tabular}
\end{table}

The difference between pixel-based and embedding-based clustering can be easily appreciated by examining the corresponding charts in Figures \ref{fig:clustering_pixel1} and \ref{fig:clustering_pixel2}. The partitions $\s{P}^g$ of embedded sequences of tiles show compact and well-separated clusters in all cases. In general, we can see that the pixel-based clusterings tend to reproduce the original images, while the clusterings based on the embedded sequences of tiles are able to disregard the specific image details to provide a more general overview of the whole region.
To support these findings, Table \ref{tab:clustering_pixel} provides numerical values for both approaches. As explained before, the Silhouette and Calinski-Harabasz scores are calculated using the distances in the two-dimensional space of the original images. This table indicates that the embedding-based clustering is clearly superior in creating dense and well-separated clusters over the geographic space.

\begin{figure*}[!t]
\centering
\includegraphics[width=1\linewidth]{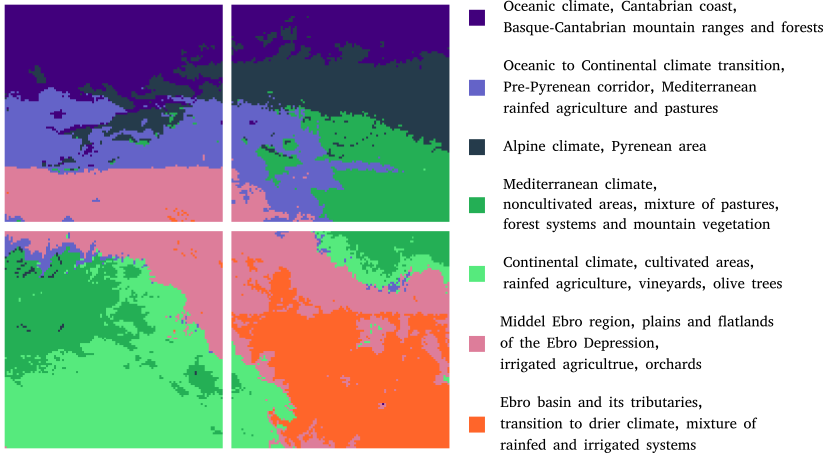}
\caption{Overall clustering and legend with associated semantic tags when considering the four sequences of images.}
\label{fig:all_k7}
\end{figure*}

To demonstrate the potential and versatility of the proposed method, we conduct the clustering of the four sequences of images as a whole. Figure \ref{fig:all_k7} demonstrates the feasibility of conducting this clustering effectively without requiring supplementary techniques like image normalization and fusion. Interconnections between the four regions naturally emerge to provide an informative partition of a large area.  
Although different clustering partitions have been explored, in Figure \ref{fig:all_k7}, we show a clustering with $K=7$. This partition into seven clusters provides the most insightful outcome, readily available for expert analysis. The legend of Figure \ref{fig:all_k7} indicates the main semantic tags linked to each cluster.
From a comprehensive standpoint, our domain expert deduces that the physical geography of this extensive region is predominantly segmented based on climatic factors. This observation further underscores the significance of incorporating time series data within the methodology.

\subsection{Clustering-based embedding}

\begin{figure*}[h!]
\centering
\subfloat[NE, $\s{P}^g$]{\includegraphics[width=0.25\linewidth]{figures/mds_NE_k5.png}}
\subfloat[NW, $\s{P}^g$]{\includegraphics[width=0.25\linewidth]{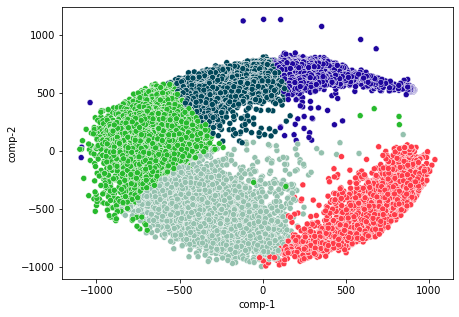}}
\subfloat[SE, $\s{P}^g$]{\includegraphics[width=0.25\linewidth]{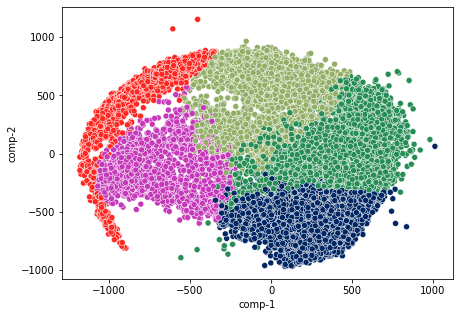}}
\subfloat[SW, $\s{P}^g$]{\includegraphics[width=0.25\linewidth]{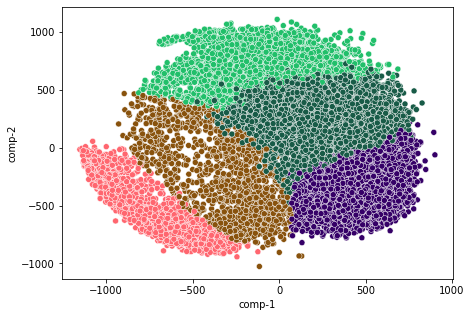}}\\
\subfloat[NE, $\s{P}^c$]{\includegraphics[width=0.25\linewidth]{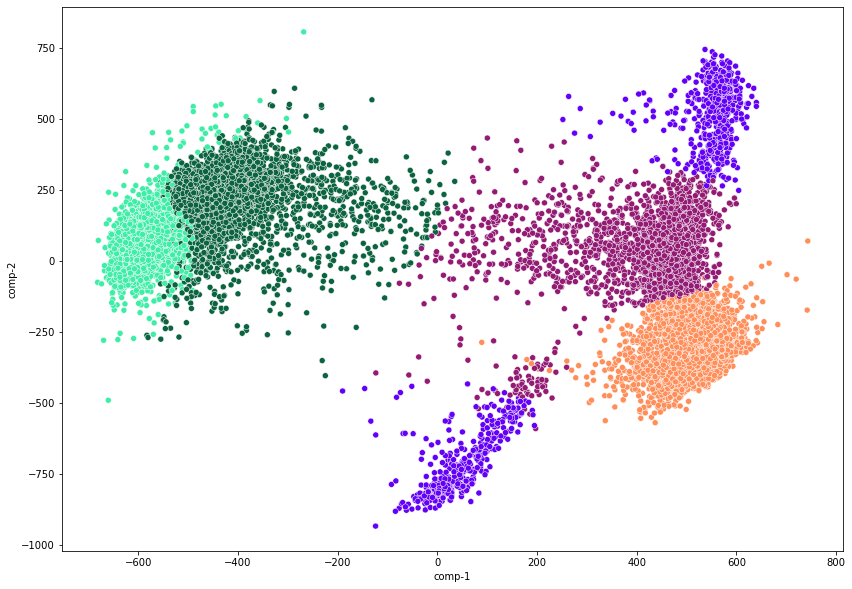}}
\subfloat[NW, $\s{P}^c$]{\includegraphics[width=0.25\linewidth]{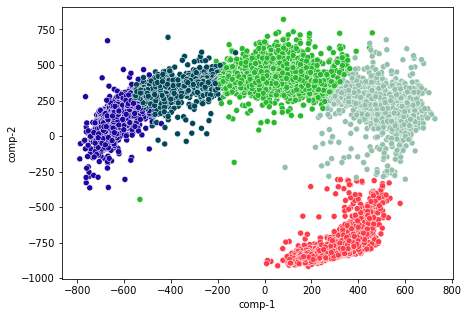}}
\subfloat[SE, $\s{P}^c$]{\includegraphics[width=0.25\linewidth]{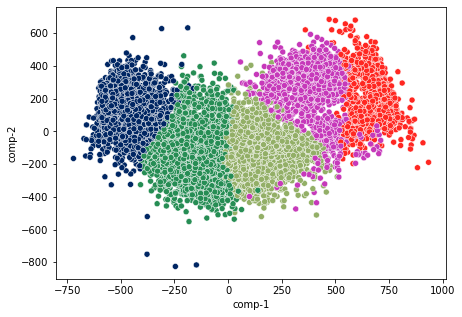}}
\subfloat[SW, $\s{P}^c$]{\includegraphics[width=0.25\linewidth]{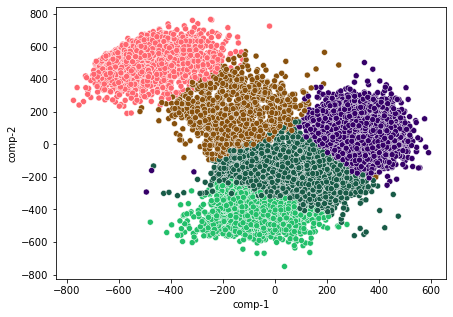}}
\caption{MDS projections, with $K=5$, of the clusterings $\s{P}^g$ and the clusterings $\s{P}^c$ for the four sequences of images.}
\label{fig:f1_f2}
\end{figure*}

This section presents the results of the clustering obtained with the refined clustering-based embedding, $f^c$. Although we select $K=5$ to train $f^c$, the impact that this second round of learning has in the embedding is independent of the number of clusters. We compare the clustering partitions $\s{P}^c$ with the preceding partitions $\s{P}^f$ for each of the four sequences, aiming to validate the method's performance.
As in the previous section, we provide both a visual comparison through the 2D projections in Figure \ref{fig:f1_f2} and a quantitative analysis using Silhouette score and the $K$-means error (intra-cluster variance) in Table \ref{tab:f1_f2}.

In Figure \ref{fig:f1_f2}, the MDS projections of the partitions $\s{P}^f$ are displayed in the first row, whereas the MDS projections of the partitions $\s{P}^c$ are shown in the second row. We chose MDS because it is deterministic and allows multiple comparisons without the need to set a random seed. For the sake of comparison, we plot the results of each region with the same colors for both kinds of partitions. The geographic representations are not shown because they are highly similar to those in Figures \ref{fig:clustering_pixel1} and \ref{fig:clustering_pixel2}, which suggests a convergence in the geographic aspect of the semantic clustering. However, if we compare the 2D projections of this figure, a dramatic change in the internal structure of the embedding is revealed in each single region. These results not only provide additional valuable information regarding the relationships among clusters but also can be useful for further automatic classification purposes or expert analysis. As evident from Table \ref{tab:f1_f2}, the clusters derived from the refined embedding exhibit greater density and enhanced separation between them, aligning with the established notion of effective clustering. Both measures indicate a higher quality of this refined clustering-based embedding for all regions. In particular, the variance within the clusters given by the $K$-means error is dramatically reduced by the refined embedding.

\begin{table}
\caption{Numerical comparison between the partitions $\s{P}^g$ and $\s{P}^c$ for $K=5$.}\label{tab:f1_f2}%
  \begin{tabular}{@{}lcccc@{}}
    \toprule
    \multirow{2}{*}{} &
      \multicolumn{2}{c}{Clustering $\s{P}^g$} &
      \multicolumn{2}{c}{Clustering $\s{P}^c$} \\
      \cmidrule(lr){2-3} \cmidrule(lr){4-5}
   Region & Silhouette & Error & Silhouette & Error \\
    \hline
    NW & 0.278 & 13458 & 0.454 & 2801\\
    NE & 0.306 & 13293 & 0.346 & 2524\\
    SW & 0.211 & 13505 & 0.329 & 2939\\
    SE & 0.203 & 14498 & 0.302 & 3222\\
    \hline
  \end{tabular}
\end{table}

These results can be combined with the previous results to achieve a better understanding of the region under study or can be used to conduct a more in-depth analysis, as we will show in the next section. For instance, Figures  \ref{fig:f1_f2} (a) and \ref{fig:f1_f2}(e) show that $\s{P}^c$ obtains a better separation between the $\wine$ and $\dgreen$ clusters than $\s{P}^g$, although both clusters are close geographically. Also note that the $\salmon$ and $\lgreen$ clusters are located on each side of the chart for $\s{P}^c$, while they are placed nearby for $\s{P}^g$. As we will explore next, the $\wine$ and $\dgreen$ clusters represent the transition towards the Pyrenees from Spain, while $\salmon$ and $\lgreen$ clusters contain different types of crops on either side of the Pyrenees. From a more global perspective, Figures \ref{fig:f1_f2} (a, e) reinforce the previous finding of two main zones in the NE region: 1) the southern zone under the Pyrenees ($\lgreen$ and $\dgreen$) and 2) the northern area in conjunction with the Pyrenees ($\salmon$, $\wine$, $\vblue$).
These observations suggest that the clustering-based embedding makes sense, and it is able to capture a better representation of the reality of the ground not captured without refinement. A similar analysis can be done in the rest of the images of Figure \ref{fig:f1_f2}. It is worth mentioning, for example, that Figures \ref{fig:f1_f2} (f) and \ref{fig:f1_f2} (g) confirm that the two clusters at the extremes of the scatter plots are indeed well separated, even in the case of Figure \ref{fig:f1_f2} (g), for which both clusters are neighbors in the geography (see Figure \ref{fig:clustering_pixel2} (i)). This figure also indicates that the most significant separation between clusters is observed in the Pyrenees.

The images in Figure \ref{fig:f1_f2} also demonstrate the convexity of the clusters formed by $K$-means over MTS. The clusters are especially well arranged with the refined embedding, where clear transitions and paths between clusters arise. Figure \ref{fig:f1_f2} (f) and \ref{fig:f1_f2} (g) arrange the clusters in a chain whereas Figure \ref{fig:f1_f2} (e) and \ref{fig:f1_f2} (h) exhibit a tree structure, where only one of the clusters has three neighbors.
These characteristics are essential to conduct a more in-depth semantic analysis in the next section, which is based on the relationship among clusters, their centroids, and the interpolations between them.

\subsection{Representatives, interpolations, and semantic tree}

\begin{figure}[!t]
	\centering
	\includegraphics[width=0.8\linewidth]{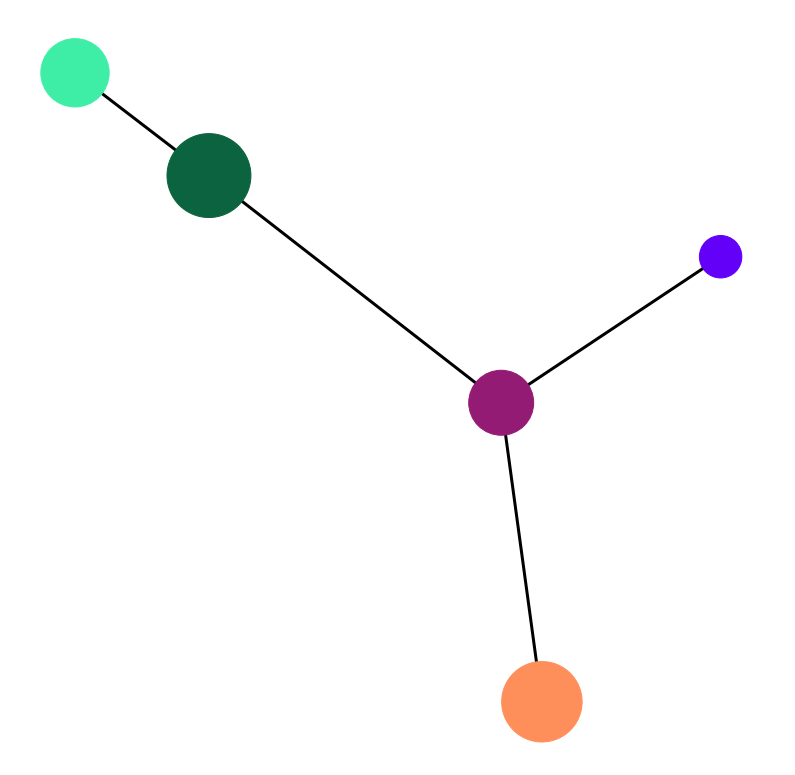}
	\caption{Semantic tree for $\mathcal{P}^c$}
	\label{fig:semantictree}
\end{figure}

In this section, we first create a semantic tree, an undirected graph that summarizes the relationships among clusters obtained from the refined embedding. As it is done in some previous experiments, we focus on the NE region to provide concise interpretations. The same analysis can be conducted for any region of interest. Moreover, to facilitate the interpretation of the figures belonging to this section, we only show the first 4 elements of the sequences of tiles that correspond to the four seasons of years 2017-2018. The semantic tree (see Section \ref{sec:tools}) of the NE region is shown in Figure \ref{fig:semantictree}. The length of the edges is proportional to the Euclidean distance between centroids and the size of the node is also proportional to the size of the cluster.

\begin{figure*}[t!]
	\centering
	\includegraphics[width=0.9\linewidth]{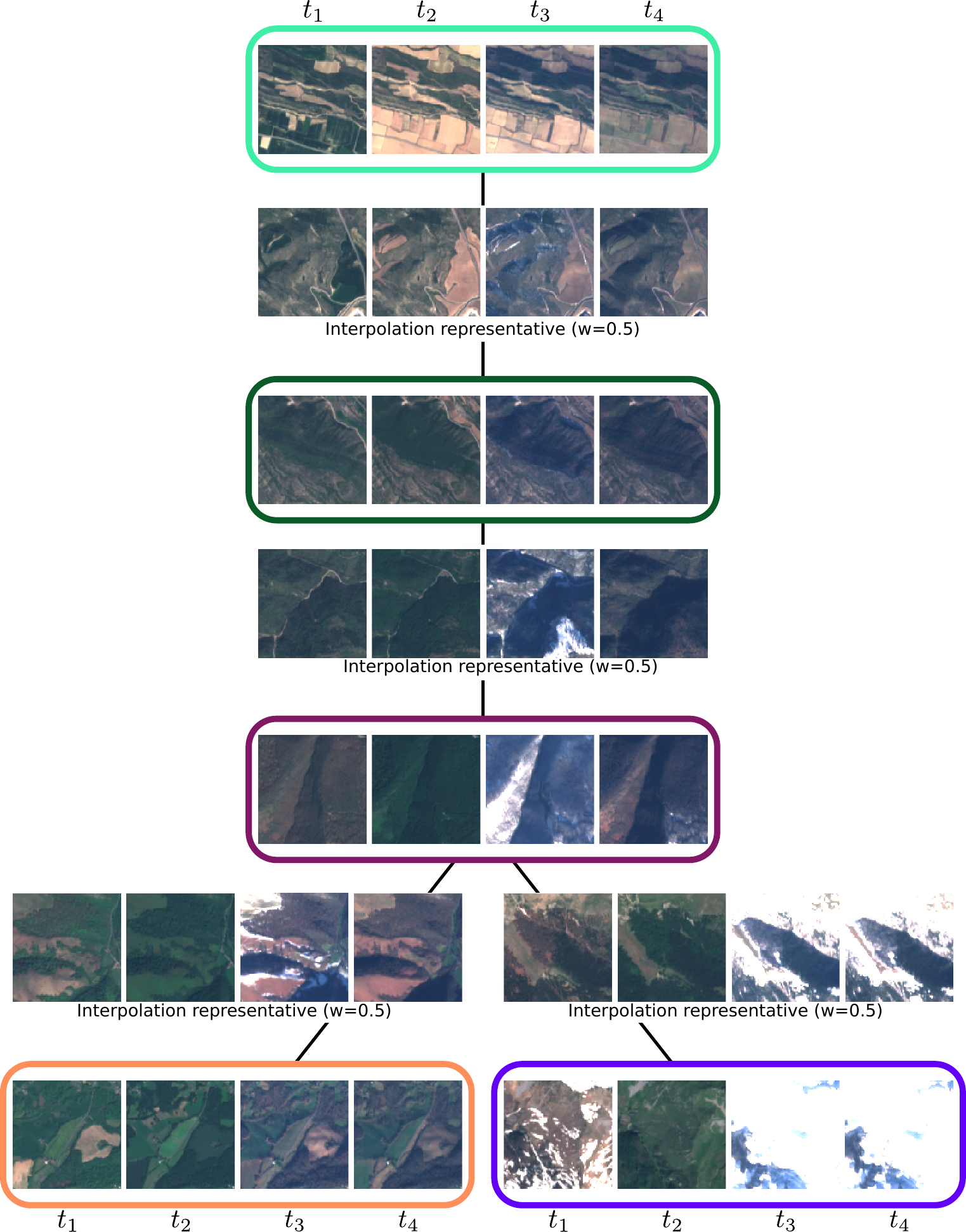}
	\caption{Semantic tree with cluster representatives and interpolations.}
	\label{fig:intertree}
\end{figure*}

Figure \ref{fig:intertree} shows the semantic tree with the representative sequence of tiles for each cluster (cluster's medoids) and the representative sequence of tiles for each edge (interpolation with $w=0.5$ between centroids). The tree representation has a unique path between every pair of nodes that exhibits a smooth evolution between their corresponding sequences of tiles.
On the one hand, Figure \ref{fig:intertree} shows that each medoid expresses a clear and well-defined overall semantic of the cluster.
Roughly speaking, the $\lgreen$ cluster is associated with crops and cultivated areas of rainfed Mediterranean agriculture. The $\dgreen$ cluster corresponds with the pre-Pyrenean area located on the southern slopes of the Pyrenees. This cluster includes valleys, hills, permanent pastures, and mountains ranging from 600 to 1800 m a.s.l. 
The $\wine$ and $\vblue$ clusters group together the Pyrenean area, where the $\vblue$ cluster encompasses the highest mountains, reaching altitudes ranging from 2000 to 2600 m a.s.l. (the maximum altitude of the Pyrenees is 3404 m a.s.l). Finally, the $\salmon$  cluster also contains a mixture of crops and pasture. However, it is located in the northern part of the Pyrenees, on the French side, where the oceanic climate prevails, leading to a different type of vegetation. Hence, we find here different crops and pastures compared to those found in cluster $\lgreen$. On the other hand, we can see that the representatives of the linear interpolations with $w=0.5$ always contain semantics halfway between the two medoids. Thus, the interpolation between $\lgreen$ and $\dgreen$ shows crops with a higher degree of pasture than the $\lgreen$ representative and includes some hills that gradually transition to low mountains in the $\dgreen$ representative. In the second interpolation, between $\dgreen$ and $\wine$, we can observe a landscape of valleys and mountains with snow in the third temporal instant as in the next node. From the $\wine$ node, we can take two paths. To the left (according to Figure \ref{fig:intertree}), we find the crops of the French side and the interpolation shows a mixture of small crops with hills and pastures. We can see that there is still some snow remaining at the third instant of time. Finally, we reach the highest peaks of the Pyrenees to the right. The interpolation exhibits a clear mountainous system that gives way to the highest and almost permanently snow-covered peaks of the $\vblue$ cluster.

\begin{figure}[t!]
	\centering
	\includegraphics[width=0.9\linewidth]{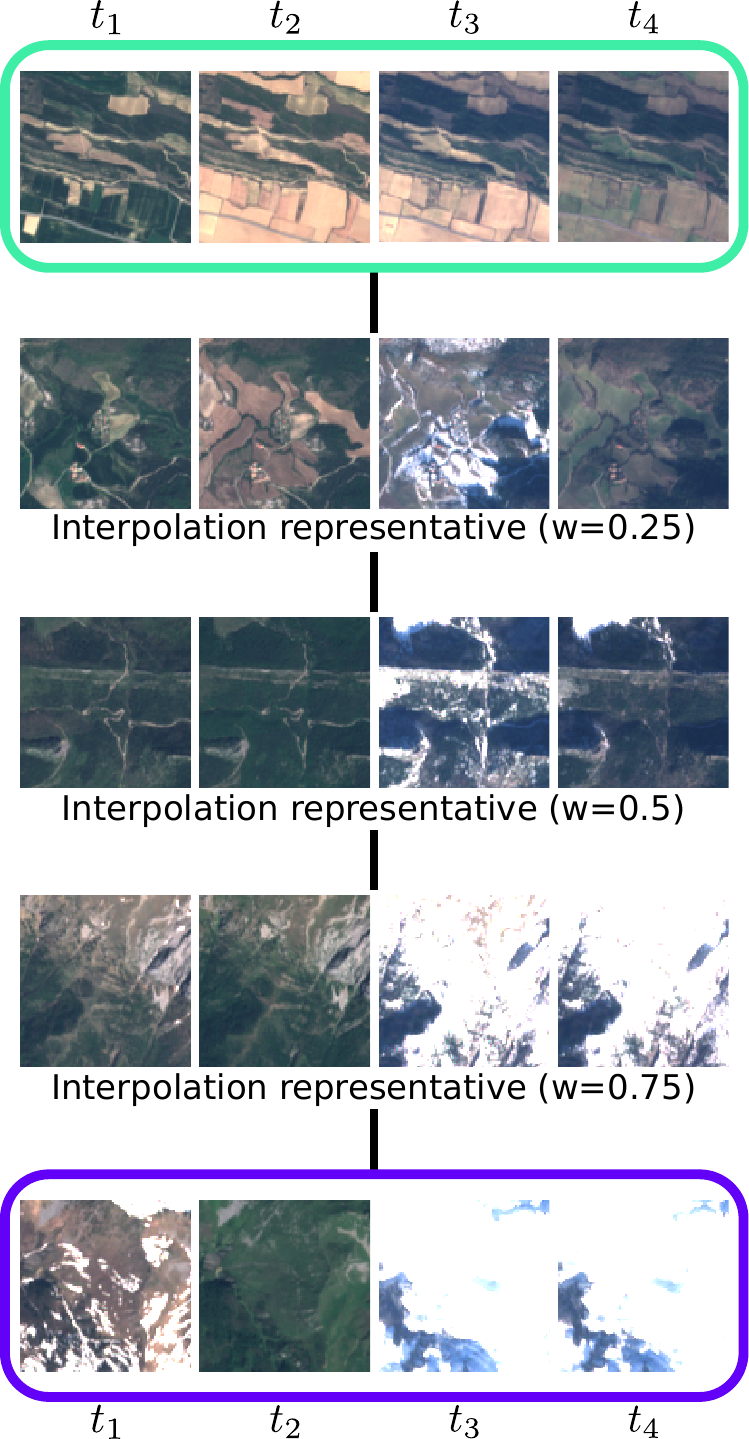}
	\caption{Interpolation between $\lgreen$ and $\vblue$ clusters in $\s{P}^c$ with $w=0.25, 0.5, 0.75$.}
	\label{fig:inter}
\end{figure}

Finally, Figure \ref{fig:inter} presents an interpolation with three intermediate steps ($w=0.25,0.5,0.75$). This figure shows a smooth transition from the $\lgreen$ to the $\vblue$ representatives. It is clearly observed that the altitude rises with increasing $w$. The land contains more pastures and mountainous terrain at each step, while the crops disappear. 
It is also worth mentioning that snow is an indicator of the different climatic patterns in each sequence of tiles. The transition is similar to that obtained in Figure \ref{fig:intertree}.

\section{Conclusions} \label{sec:conclu}

In this paper, we investigate a fully unsupervised methodology to conduct semantic clustering of a large region using sequences of satellite images. The sequences of images are encoded as a set of multidimensional time series (MTS) using a semantically meaningful embedding, which is built in three main steps: 1) training the embedding with triplets generated according to the geographic neighborhood, 2) clustering the MTS, and 3) embedding refining with triplets generated according to the clustering neighborhood. 

We conduct experiments to explore the clustering partitions, thereby acquiring valuable comprehensive insights into the region. We observe that the geographic representation of the clustering reveals a distinct structure, wherein extensive regions are tightly grouped compactly. Moreover, as the number of clusters increases, a hierarchical partition naturally emerges, revealing an increasing number of semantic details, that could be studied more in-depth depending on the specific application. Both the geographical and embedded representations offer supplementary information regarding the configuration of the clusters and the interconnections between them.
In particular, the refined clustering-based embedding is able to sharpen the semantic information obtained from the sequence of satellite images. This embedding exploits the information about the underlying properties of the land for a given number of clusters. The clustering-based embedding highlights the elements belonging to each cluster, i.e., it tries to separate the borders and brings the points to the center of the cluster, which is desirable for further interpolation analysis or classification tasks.
The visual inspection of the centroids and the corresponding interpolations show the ability of the clustering to capture the different semantics of a region and its evolution. Each cluster is able to represent a specific and well-differentiated spatio-temporal semantic. Also, the interpolation between clusters is clearly meaningful due to the convex partitions provided by $K$-means and the impact of the refined embedding.

 The proposed procedure is applied to a region of northern Spain, but any other region of interest may be studied. The results of the method do not reproduce very specific details of the images as pixel-based partitions do but provide a higher level of abstraction embodied in a coarse-grain clustering approach. This semantic arrangement of the land is particularly distinguished by finding structured patterns, where the clusters tend to cover wide and compact areas that put together complex evolving semantics such as the Ebro river basin or the Pyrenees and separate climatic zones, for instance, clearly distinguishing between oceanic and continental climates. To the best of our knowledge, this kind of high-level semantic partition has not been carried out automatically as a fully unsupervised procedure before.
The proposed methodology can be used to assist non-technical users, such as decision-makers, domain experts, or policy-makers, to easily gain insights from the data without the need for extensive data processing, technical expertise, or labeling efforts. Thus, the provided visualizations, succinct summaries, and practical information can be readily understood at user-level.

In future work, we will continue developing the methodology and exploring its applicability to various challenges, including climate zone mapping, river basin analysis, or assessment of ecological and agricultural systems. We argue that our methodology is also suitable to analyze critical environmental issues such as desertification.
This can help to encode indicative spatio-temporal patterns and identify affected areas, providing valuable insights for studies and mitigation measures. Furthermore,  the flexibility of the clustering procedure proposed in this work could be used in distributed computing as it
allows for the division of large spatial regions, which facilitates the development of strategies to allocate similar areas to different machines.
One significant drawback of distributed processing is the slowness in transferring information between machines. This spatio-temporal clustering approach has the potential to mitigate these limitations by grouping information both in spatial and temporal dimensions. Additionally, since the clusters are interpretable, it becomes feasible to identify both nearby and distant clusters, allowing for flexible and automatic changes in the groupings. This behavior would enable an elastic division of the processing system, defining how to distribute similar data on a new machine if necessary.

Finally, it is important to remark that using sequences of tiles is an essential element in conducting further analysis related to the changing semantics of a region. The specific use of sequences of tiles incorporates a new dimension in semantic clustering that provides richer information and presents a wide variety of possibilities. For instance, bi-clustering algorithms can be used with the set of MTS to find similar sub-sets both in space and time, detecting change points and seasonality in land semantics.  

\section*{Acknowledgments}
We express our gratitude to Juan José Calvo, an
expert geographer from the Public University of Navarre with in-
depth knowledge of the study region, for validating our experimental
findings. This work has been supported by Project PID2020-113125RB-
I00/MCIN/AEI/10.130 39/501100011033. Aritz Pérez has been sup-
ported by Basque Government through the Elkartek program and the
BERC 2022-2025 program, and by the Ministry of Science and Innova-
tion: BCAM Severo Ochoa accreditation CEX2021-001142-S/ MICIN/
AEI/ 10.13039/ 501100011033.

\bibliography{biblio.bib}

\end{document}